\PassOptionsToPackage{table,svgnames}{xcolor}

\documentclass[pmlr,twocolumn,10pt]{jmlr} 
\hypersetup{
    citecolor=NavyBlue,
    linkcolor=NavyBlue,
    urlcolor=NavyBlue
}
\usepackage{float}

\usepackage{booktabs}
\usepackage{siunitx}

\usepackage[switch]{lineno}
\usepackage{soul}

\usepackage[nameinlink,capitalize,noabbrev]{cleveref}

\theorembodyfont{\upshape}
\theoremheaderfont{\scshape}
\theorempostheader{:}
\theoremsep{\newline}

\jmlryear{2026}

\title[ML-AUTOC]{Precedence-Aware Resource Allocation: \\Extending the AUTOC Framework to Multi-Level Treatments}

\author{%
\Name{Olin E. Gilster}
\\
\addr Work done while at Emory University, USA
\AND
\Name{Shengpu Tang} \Email{shengpu.tang@emory.edu}\\
\addr Emory University, USA
\vspace{-1em}
}

\begin{document}

\maketitle

\begin{abstract}
Many real-world decision problems require allocating limited resources across individuals to maximize total benefit often through interventions that vary in intensity rather than binary treat-or-not decisions. In these settings, higher levels treatments require first applying lower levels of treatment, inducing a precedence constraint that complicates allocation under a fixed budget. While the causal inference literature has largely emphasized accurate estimation of conditional average treatment effects (CATE), recent work has shown that estimation accuracy alone does not guarantee effective resource allocation, motivating ranking-based evaluation frameworks. The Area Under the Targeting Operating Characteristic (AUTOC) provides an effective way to evaluate prioritization rules across budget levels, but existing formulations are limited to binary treatments. We introduce Multi-Level AUTOC (ML-AUTOC), which is a generalization for AUTOC designed for multi-level treatment settings with precedence. We also propose precedence-aware policies for mapping true marginal treatment effects to feasible treatment assignments under fixed budget constraints. Our results show that in the setting of multi-level treatments, maximizing realized treatment effects may require re-allocation of treatments across individuals. 
\end{abstract}

\paragraph*{Data and Code Availability.}
All code used to generate the experiments and figures in this paper, along with supporting documentation, is publicly available at \url{https://github.com/gilstero/ML-TOC}. All datasets used in this study are synthetically generated and can be reproduced using the provided data generation scripts.

\paragraph*{Institutional Review Board (IRB).}
This study does not require IRB approval.

\vspace{-0.5em}

\section{Introduction}
\label{sec:intro}
\vspace{-0.5em}

Across the domains of healthcare, education, and public policy, decision makers are often faced with the reality of resource constraints. For example, hospitals may find themselves confronting a surge of critically ill patients with limited number of beds or personnel. These constraints force hospitals to allocate their resources based on need, deciding who gets intensive care versus standard or minimal care \citep{french2021hospitalstrain}. Beyond healthcare, other domains also face similar challenges. In education, overcrowded schools may limit the amount of attention each child gets. Thus, school systems need to make informed decisions on who to assign excess teacher hours to maximize school test scores and student well-being \citep{vakili2024overcrowded}. In the public-policy setting, limited staffing at social outreach programs can reduce the amount of care given to each person. Specifically, in the context of housing, people may be assigned to increasingly better programs from basic shelter, transitional housing, and long-term assistance fully based on the amount of resources available to an organization \citep{agonafer2021community}.

The stakes of such allocation decisions are particularly acute in the healthcare setting \citep{liu2013hospital}. Recent papers covering the COVID-19 pandemic found that when hospitals experienced a capacity strain, they had higher rates of mortality. Specifically, they found that in a nationwide resource constraint where ICU capacity was at 75\%, “12,000 excess deaths would occur in the following two weeks." When ICU capacity was at 100\%, it was predicted that 80,000 excess deaths would occur in the same time period \citep{french2021hospitalstrain}. Together, these examples highlight the widespread challenge of allocating limited resources and the high cost of misallocation, emphasizing the need for systematic approaches to improve decision-making under resource constraints.

Researchers working at the intersection of causal inference and machine learning have largely focused on the problem of conditional average treatment effect (CATE) estimation, aiming to minimize the error between predicted and true individual treatment effects \citep{shalit2017estimating,kunzel2019metalearners}. 
Yet, accurate CATE estimation does not necessarily translate to optimal resource allocation. 
Thus, researchers have increasingly pivoted toward ranking-based optimization and evaluation.
For instance, \citet{kamran2024learning} proposed algorithms that directly optimize rankings for expected benefit, while \citet{caniglia2021estimating} utilized dynamic marginal structural models (MSMs) to identify optimal strategies under resource usage constraints. 
To assess the quality of prioritization rules independently of estimation accuracy, \citet{yadlowsky2025evaluating} popularized the area under the targeting operating characteristic (AUTOC) metric, previously described in \citet{zhao2013effectively}. 
This metric quantifies how effectively a scoring function prioritizes individuals compared to a random allocation baseline across all possible budget levels. 

Despite the utility of the AUTOC metric, existing works on ranking-based evaluation frameworks are limited to binary treatment decisions, where each individual is either treated or not treated. Current frameworks do not adequately address settings where interventions vary in intensity and follow a precedence structure, where higher-intensity treatments require first receiving lower-intensity ones. As our paper will illustrate, naively converting such problems into a binary treatment formulation obscures these nuances and can lead to suboptimal allocation decisions. Therefore, extending ranking-based allocation rules and evaluation beyond the binary treatment setting is an essential building block for applying them to a larger range of real-world decision problems.

In this work, we propose a generalization of the binary AUTOC metric to settings involving multi-level treatments. 
This extension introduces unique challenges, primarily due to the requirement of treatment level precedence. 
While a scoring function in a binary treatment setting maps to a unique ranking of individuals, our setting utilizes scoring functions that estimate marginal treatment effects for each individual at each treatment level. 
Every scoring function may correspond to multiple possible rankings of individual–level pairs, some respecting precedence, and some that do not. 
Furthermore, the random allocation baseline must also be redefined to respect precedence. 
Finally, as resource budget increases, the optimal allocation may require re-allocating previously assigned resources to different individual–level pairs. This leads to the phenomenon of ``flipping'' or non-monotonic assignment, which does not exist for binary treatment settings by definition. 
Specifically, this paper makes the following contributions: 
\begin{itemize}
\setlength{\topsep}{0pt}
\setlength{\itemsep}{0pt}
    \item We develop and formalize the multi-level AUTOC (ML-AUTOC) metric to evaluate prioritization efficiency of resource allocation tasks with multiple treatment levels.
    \item We introduce and compare several resource allocation rules, which we refer to as policies, that can be derived from scoring functions, and evaluate their ability to maximize realized treatment effects on synthetic datasets.
    \item We show that in multi-level treatment settings, the re-allocating \textsf{PAO} policy fundamentally departs from ranking-based approaches, since maximizing realized treatment effects may require non-nested allocation policies that switch between partially treating one individual and initiating treatment for another.
\end{itemize}

\subsection{Related Work: CATE}

To understand the motivation for AUTOC and our extension to multi-level settings, it is important to examine the limitations of conditional average treatment effect (CATE) estimation in resource constrained problems. 
Consider the dataset \(S={(x_i}, t_i,y_i)^n_{i=1}\), where each individual $i$ has covariates $x_i\in X \subseteq \mathbb{R}^d$, assigned treatment $t_i \in \{0,1\}$, and observed outcome $y_i \in \mathbb{R}$. Under the potential outcomes framework \citep{Rubin1974}, each individual is given potential outcomes $Y_i(0)$ and $Y_i(1)$ corresponding to control and treatment. Given these, CATE is defined as: 
\[
\tau_i = \mathbb{E}[Y_i(1) - Y_i(0) \mid x_i].
\]
CATE allows researchers to go beyond a single global estimate of the average treatment effect (ATE) by estimating $\tau_i$ conditioned on an individual's covariates $x_i$ \citep{kunzel2019metalearners}. Despite the success in CATE research, methods often focus on estimation accuracy rather than allocation efficiency \citep{kamran2024learning, yadlowsky2025evaluating}. While CATE-based models can estimate treatment effect heterogeneity with high precision, they may still lead to suboptimal allocation decisions when resources are limited. It was found in \citet{kamran2024learning} that minimizing CATE predictive error does not necessarily translate to maximizing total utility under a budget constraint. Moreover, most CATE frameworks implicitly assume that treatment can be assigned without accounting for constraints on total resource use \citep{caniglia2021estimating}. In practice, interventions often vary in cost and level, which require trade-offs where only a subset of individuals can be treated under a given budget. These limitations have motivated the development of ranking-based frameworks, such as AUTOC, which explicitly evaluates how effectively a model prioritizes individuals for treatment as available resources vary \citep{kamran2024learning, yadlowsky2025evaluating}. 

\subsection{Related Work: AUTOC}

AUTOC provides the basis for evaluating treatment prioritization when the treatment structure is binary, and ML-AUTOC closely mirrors this framework in the multi-level setting; for this reason, it is important to first understand AUTOC.

The goal of AUTOC is to evaluate how well a scoring function $f(x_i)$, which can be mapped to a ranking over individuals, aligns with the true ranking based on CATE $\tau_i$. Given a treatment budget $u \in [0,1]$, $D_u^S(f)$ denotes the top $u$-fraction of individuals in $S$ ranked the highest by score $f$:
\[
D_u^S(f)=\{i \mid f(x_i)\geq \psi(f(x_i)_{i\in S}, u)\}
\]
where $\psi(\cdot, u)$ represents the $(1-u)$ quantile of model scores. The ATE at budget $u$ induced by the scoring function $f$ is defined as the mean of individual treatment effects for those individuals selected to be treated:
\[
\text{ATE}_u^S(f)=\frac{1}{D_u^S(f)}\sum_{i\in D_u^S(f)} \tau_i.
\]

The targeting operating characteristic (TOC) at $u$ quantifies how much better this ATE is than that of a random selection at the same budget. In the binary case, it is straightforward to obtain the random baseline since
the expected ATE under random assignment is equal to the population ATE. Thus, the TOC is defined as:
\[
\text{TOC}_u^S(f)=\text{ATE}_u^S(f)-\frac{1}{|S|}\sum^{|S|}_{i=1}\tau_i
\]
Finally, the area under the TOC curve averages over all budgets:
\[
\text{AUTOC}_S(f)=\frac{1}{|S|}\sum^{|S|}_{i=1}\text{TOC}^S_{\frac{i}{|S|}}(f).
\]
Intuitively, AUTOC captures the average improvement in treatment benefit obtained by using the model to evaluate patients at every possible resource level. A larger AUTOC value indicates that a ranking is more accurately prioritizing individuals who would benefit most from treatment, while an $\text{AUTOC} = 0$ is equivalent to random allocation \citep{kamran2024learning, yadlowsky2025evaluating}. 

\section{Methods}

\subsection{Problem Statement}

While AUTOC provides a strong evaluation metric for binary treatment allocation \citep{french2021hospitalstrain,vakili2024overcrowded,agonafer2021community}, many decision problems fall into \textit{multi-level} intervention decisions where treatment intensity can be quantified as discrete levels $\ell \in \{0,1,\dots L\}$. In our case, these treatment levels exhibit precedence; that is, higher treatment levels may only be assigned if previous levels have been assigned. Each higher level typically provides a positive marginal benefit, denoted:
\begin{equation}
\delta_i^{(\ell)} = \mathbb{E}[Y_i(\ell) - Y_i(\ell-1) \mid x_i]\label{eq:1}   
\end{equation}
representing the incremental gain from advancing an individual $i$ from level $\ell-1$ to $\ell$. Standard AUTOC \citep{yadlowsky2025evaluating,kamran2024learning,zhao2013effectively} which assumes a binary decision space $t_i\in\{0,1\}$ cannot encapsulate these hierarchical relationships because it evaluates rankings on a single treatment basis.


Consider the dataset $S={(x_i}, t_i,y_i)^n_{i=1}$ where each individual $i$ has covariates $x_i \in X$, assigned treatment levels $t_i\in \{0,1,\dots L\}$, and observed outcome $y_i$. Because of the treatment structure, each individual has a vector of potential outcomes: 
\[
Y_i = (Y_i(0), Y_i(1),\dots,Y_i(L)),
\]
which can be derived from the marginal treatment effects in \cref{eq:1}. Given our goal is to evaluate rather than optimize a scoring function, in our experiments, we do not model or simulate covariates $X$. Instead, we directly simulate the outputs of the scoring functions in order to study how a given ranking should be translated into a treatment allocation policy under a fixed budget.

There remains a challenge in defining the budget for ML-AUTOC. In the standard AUTOC metric, the unit for each incremental movement in the budget is the choice to treat an individual. In our setting, individuals can be treated multiple times, which invalidates using them as a whole unit. To address this, we define the budget in terms of incremental treatment steps. Moving an individual from treatment level \(\ell-1\) to level $\ell$ consumes one unit of budget, providing a natural and consistent unit of resource expenditure in the multi-level setting. One may alternatively assume that higher treatment levels cost more (or less) than lower treatment levels, butu for simplicity, we focus our discussion within the unit-cost setting. 

Our problem setting is closely related to multi-period precedence-constrained knapsack problems, which study the allocation of resources across hierarchical or ordered choices under a fixed budget. Prior work has explored such problems in static or multi-period settings (e.g., \citet{moreno, ASLAN2023102779, SAMAVATI201712}), typically focusing on optimizing a single objective value at a fixed budget.

\subsection{Solutions via Dynamic Programming}

The allocation problem that underlies ML-AUTOC is closely related to precedence-constrained multiple-choice knapsack problem (MCKP). Each individual $i$ represents a class of mutually exclusive cumulative treatments, corresponding to levels $\ell \in \{0,\dots,L\}$, and each level produces cumulative benefit
\[
V_i(\ell) = \sum_{k=1}^{\ell} \delta_i^{(k)} = \mathbb{E}[Y_i(\ell) - Y_i(0) \mid x_i].
\]
The planner must choose the level of treatment per individual such that the total budget does not exceed $\beta$ and the total benefit is maximized. 
Formally, the problem for a fixed budget $\beta$ is:
\[
\max_{\ell_i \in \{0,\dots,L\}} \sum_{i=1}^{n} V_i(\ell_i)
\quad \text{subject to} \quad
\sum_{i=1}^{n} \ell_i \leq \beta.
\]
This formulation is a special case of the MCKP, one of the most well-studied combinatorial optimization problems \citep{freville2004multidimensional, KellererPferschyPisinger2004}. While integer linear programming (ILP) formulations are common, dynamic programming (DP) methods provide a natural and computationally efficient approach when budgets are discrete and relatively small \citep{MartelloToth1990}.

Dynamic programming is particularly well suited to our setting for two reasons:
\begin{enumerate}
\setlength{\topsep}{0pt}
\setlength{\itemsep}{0pt}
    \item \textbf{Evaluation Across All Budgets.} ML-AUTOC requires evaluating optimal allocations for every $\beta \in \{0,1,\dots,nL\}$. A single DP table computes optimal values for all budgets simultaneously, whereas solving separate ILPs for each $\beta$ would be redundant.
    \item \textbf{Ranking Recovery.} The DP solution naturally stores dependencies between solutions to sub-problems of different $\beta$ values. By tracing which individual–level pair (marginal treatment) is selected at each increment of $\beta$, we recover an implied ordering of treatment increments.
\end{enumerate}
More information on the exact dynamic programming implementation can be found in \cref{apd:dynamic}.

\subsection{Scoring Functions, Rankings, and Allocation Policies}



Before describing our proposed approach, we need to distinguish a few related concepts that are equivalent under the binary treatment setting, but becomes tricky under multi-level treatments. In the binary treatment setting, a scoring function induces a unique ranking of individuals (up to tie-breaking), and treating the top ranked individuals at each budget defines an allocation policy. As a result, scoring functions, rankings, and allocation policies are all effectively interchangeable. 
In the multi-level setting, this equivalence no longer holds. A multi-level scoring function associates each treatment level $\ell$ with a level-specific scoring function $f_\ell(x_i)$. These scores are intended to rank individuals according to their predicted marginal benefit at level $\ell$, $\delta_i^{(\ell)}$. Thus, the scoring function assigns values to individual–level pairs rather than individuals. Unlike in the binary case, these level-specific scores by themselves do not define how treatments should be allocated across levels and individuals as the budget increases. 
From the scoring function, we construct a policy $\pi$ that selects a set of individual–level pairs $(i,\ell)$ subject to a budget constraint. Let
\[\Delta = \{(i,\ell) : i \in \{1,\dots,n\},\; \ell \in \{1,\dots,L\}\}\]
denote the set of all individual–level pairs. We define a policy as a mapping from a budget level $\beta \in \{0,1,\dots,nL\}$ to a feasible subset of pairs:
\[\pi : \{0,1,\dots,nL\} \rightarrow 2^\Delta, \quad \beta \mapsto \pi(\beta),\]
where $\pi(\beta) \subseteq \Delta$ contains the pairs selected by $\pi$ under budget $\beta$, and precedence constraints restrict what treatment allocations are feasible at a given budget. As the budget increases, the sequence of sets $\{\pi(\beta)\}_\beta$ (possibly) induces an ordering over treatment assignments. We interpret this sequence as a ranking over feasible individual–level pairs. 


The best way to conceptualize the two objects, \(f_\ell(x_i)\) and \(\pi\), is to consider the scoring function first. The scoring function produces a level-specific ranking of treatment assignments based on predicted marginal benefits, which can be interpreted as a priority list. A policy then takes these rankings and applies a set of allocation rules, such as budget constraints, precedence requirements, or aggregation across levels, to determine which individual–level pairs are ultimately selected. Importantly, this mapping from \(f_\ell(x_i)\) to \(\pi\) is not unique, unlike in the binary treatment setting. As we will demonstrate later in this section, it is possible that the same scoring function may give rise to multiple feasible policies, and some allocation policies can not be represented as a single global ranking. 

Consider the following toy example in \cref{tab:1}:

\begin{table}[ht]
\vspace{-1.5em}
    \centering
    \caption{True Marginal Benefit table}
    \label{tab:1}
    \begin{tabular}{|c|c|c|} \hline
    $\delta_i^{(\ell)}$ & $\ell = 1$ & $\ell = 2$ \\ \hline
    $x_1$ & 1 & 5 \\ \hline
    $x_2$ & 2 & 3 \\ \hline
    \end{tabular}
\end{table}

\vspace{-0.75em}

At level $\ell = 1$, since $\delta_2^{(1)} > \delta_1^{(1)}$, individual $x_2$ should receive priority for the first unit of treatment. However, at level $\ell = 2$, the ordering reverses, as $\delta_2^{(2)} < \delta_1^{(2)}$. This highlights a key challenge in the multi-level setting: \textbf{a single global score cannot, in general, correctly prioritize treatment assignments over individuals across all levels.}
For instance, a global scoring rule of the form $f(x_i) \propto Y_i(L) - Y_i(0)$ (or equivalently $\sum_{\ell=1}^{L} \delta_i^{(\ell)}$) would rank $x_1 \succ x_2$ overall in this example ($6 > 5$). Under tight budgets, such a ranking would incorrectly allocate the first unit of treatment to $x_1$, despite $x_2$ having the larger marginal benefit at $\ell = 1$.

\subsection{Naive Policies}

\vspace{-0.5em}
To map a scoring function to a policy in the multi-level treatment setting, one could naively convert the problem to a binary treatment problem. Below, we discuss three possibilities:

\begin{itemize}
\setlength{\topsep}{0pt}
\setlength{\itemsep}{0pt}
    \item Pooled Global Policy (\textsf{PGP}): \\
    This policy creates a global ranking by pooling all marginal scores $\{f_\ell(x_i) : \forall i, \ell\}$ across individuals $i$ and levels $\ell$, and selecting the highest values according to the marginal score ranking (e.g. 5, 3, 2, 1). 
    While this identifies high-benefit individual–level pairs, it frequently violates precedence constraints, for example, assigning Level 2 to an individual before they receive Level 1. 

    \item Individual-First Policy (\textsf{IFP}) \\
    This approach aggregates each individual's total potential benefit, $\sum_\ell f_\ell(x_i)$, and ranks all individuals by this aggregate score (e.g., $1+5=6$, $2+3=5$). Resources are allocated to the top-ranked individual until they are fully treated across all levels before moving to the next individual (e.g., 1, 5, 2, 3). Within each individual, treatment levels are assigned in increasing order of intensity, enforcing the precedence-induced ranking of levels. While this respects precedence, it cannot spread lower-level treatments across many individuals, even when the marginal benefit of Level 1 is significantly higher than the marginal benefit of Level 2.

    \item Level-First Policy (\textsf{LFP}): This policy imposes a strict hierarchy based on treatment levels, similar to classical level-based scheduling algorithms \citep{6c502739-05f5-3827-8ddf-208f1e1191cf}. It requires that every individual in the population receive Level 1 before any individual can be considered for Level 2. Within each level, individuals are ranked by the marginal benefits of that level, e.g., allocating Level 1 in the order (2, 1) and allocating Level 2 in the order (5, 3) for the overall ranking of (2, 1, 5, 3).
\end{itemize}




\subsection{Precedence-Aware Policies} \label{section 2.6}

Unfortunately, these rankings fall short as no specific one fully captures the structure of the problem: they ignore how marginal benefits vary by level or fail to respect the hierarchical dependencies across treatment levels. What we want is a ranking that i) accounts for level-specific marginal benefits and ii) preserves precedence constraints. 
Because the \textsf{PGP}, \textsf{IFP}, and \textsf{LFP} policies fail to balance precedence with marginal utility, we propose two precedence-aware policies that dynamically navigate these trade-offs to maximize total population benefit.

A feasible allocation must satisfy the precedence constraint that, for any individual $i$, treatment level $\ell$ can only be assigned if all lower levels have already been assigned. Formally, if $(i,\ell)$ is selected, then $(i,k)$ must also be selected for all $k < \ell$. Once taking into account what possible treatments are available at a given time, we use a greedy strategy to select the one with the highest marginal benefit as the next one to treat. Two possible policies can be built off of these rules, which we call Precedence-Aware Greedy (\textsf{PAG}) and Precedence-Aware Optimal (\textsf{PAO}). The distinction between the two rankings lies in the reallocation of treatments at any distinct budget. The \textsf{PAG} policy constructs the ranking incrementally, fixing earlier treatment assignments and never revising past decisions as the budget expands. In contrast, the \textsf{PAO} policy allows re-optimization at each budget level, potentially revising earlier allocations to account for newly feasible treatment options. From an algorithmic perspective, \textsf{PAO} corresponds to solving a sequence of budget-constrained optimization problems via dynamic programming. Although we do not discuss it here, a detailed description of the dynamic programming formulation and its computational complexity is provided in \cref{apd:dynamic}.

For the toy example in \cref{tab:1}, each policy is evaluated compared to a random baseline allocation under the same budget (defined in \cref{2.7}). In this specific case, all feasible sets were calculated to determine a true average for the random baseline. We plot the TOC curves for each policy, showing budget levels versus average benefit across the cohort:

\begin{figure} [ht]
    \centering
    \caption{Toy example ML-TOC Graph}
    \includegraphics[width=0.9\linewidth]{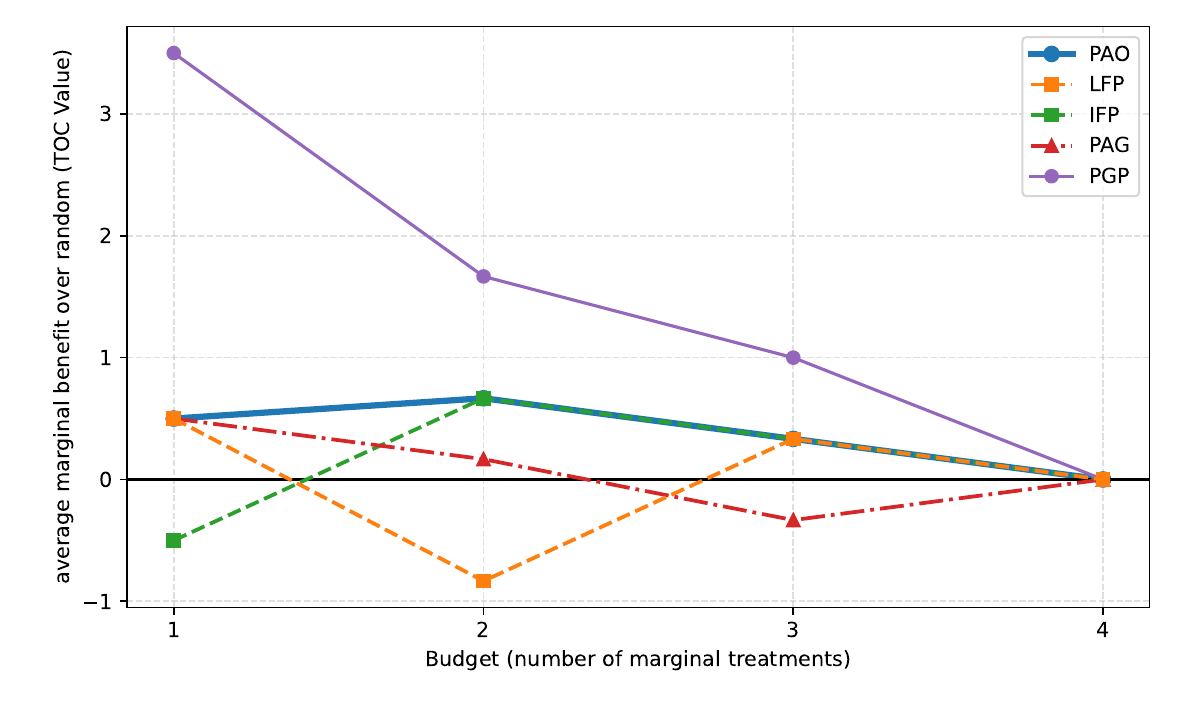}
    \label{fig:1}
\end{figure}

The TOC curves in \cref{fig:1} illustrate how different policies behave under the same non-binary marginal benefit structure. The \textsf{PAO} policy achieves the greatest benefit while adhering to precedence constraints. The \textsf{PAG} policy matches \textsf{PAO} at the smallest budget but falls behind at intermediate budgets, since its nested allocation cannot revise earlier choices, while \textsf{IFP} matches \textsf{PAO} from the second budget onward. The \textsf{PGP} appears to perform best, but its advantage is artificial, as it violates precedence twice. This illustrates why multi-level rankings cannot be evaluated correctly if these constraints are not enforced.

\subsection{Formalizing ML-AUTOC} \label{2.7}

Because individuals can be treated incrementally across multiple levels, defining treatment thresholds in terms of the top $u\%$ of individuals is not a meaningful metric. Instead, we index allocations by a scalar budget parameter $\beta \in \{1,\dots,nL\} = \mathcal{B}$, which represents the total number of treatment units assigned across all individuals and levels. For a given scoring function $f = \{f_\ell : \ell = 1 \cdots L\}$, we define the allocation set under budget $\beta$ as:
\begin{equation}
    D_{\beta}^S(f) = \pi_f(\beta)
\end{equation}
where $\pi_f(\beta) \subseteq \Delta$ is the set of individual–level pairs selected by the policy induced by $f$ under budget $\beta$. The average marginal benefit realized under budget $\beta$ is then:
\begin{equation}
    \text{ATE}_{\beta}^S(f) =
    \frac{1}{|D_{\beta}^S(f)|}
    \sum_{(i,\ell) \in D_{\beta}^S(f)} \delta_i^{(\ell)}.
    \label{eq:3}
\end{equation}
The targeting operating characteristic at budget $\beta$ quantifies how much better the selected subset is compared to a baseline equivalent to random selection and needs to respect the treatment level precedence structures. 
To calculate the random baseline, we define $\mathcal{F}_\beta$ as the set of all feasible allocations of individual–level pairs under a budget $\beta$ that satisfy precedence. We consider a random allocation $\widetilde{A}_\beta$ drawn uniformly from $\mathcal{F}_\beta$. Therefore, for all $(i, \ell) \in \widetilde{A}_\beta$ we can define $\text{RAND}_\beta$ as the $\mathbb{E}_{\widetilde{A}_\beta}[\sum_{(i,\ell)\in \widetilde{A}_\beta}\delta_i^{(\ell)}]$. In practice, this can be calculated via enumeration (for small $n$ and $L$) or Monte Carlo sampling to approximate the expectation. Now for a scoring function that induces a feasible allocation, TOC can be defined as: 
\begin{equation}
    \text{TOC}_{\beta}^S(f)
    =
    \text{ATE}_{\beta}^S(f)
    -
    \text{RAND}_\beta.
    \label{eq:4}
\end{equation}
Let $\mathcal{B}$ denote the set of feasible budgets, with $|\mathcal{B}|$ its cardinality, so that $\frac{1}{|\mathcal{B}|}$ averages over these budgets. Finally, the ML-AUTOC score aggregates realized average treatment effects across all budgets:
\begin{equation}
    \text{ML-AUTOC}_S(f)
    =
    \frac{1}{|\mathcal{B}|}
    \sum_{\beta \in \mathcal{B}}
    \text{TOC}_{\beta}^S(f).
    \label{eq:5}
\end{equation}
ML-AUTOC and AUTOC are designed to measure the same conceptual object, which is how well a learned scoring rule prioritizes units for treatment as resources increase. The difference is in how the comparison is made. AUTOC is defined for binary treatment, where each additional unit of budget corresponds to treating one new individual and the random baseline is a uniform draw over subsets of individuals of a given size. In ML-AUTOC, the budget expands over treatment increments $(i,\ell)$ which are subject to precedence, so the random baseline must be defined over the feasible allocation set. This change in the feasible set (and therefore in the baseline being subtracted) is the fundamental difference between AUTOC and ML-AUTOC.

A second difference is the unit for which the area is accumulated. In binary AUTOC, the horizontal axis is naturally interpreted as the treated fraction $u$. In the multi-level setting, the natural budget unit is the number of incremental level assignments, so the curve aggregates performance over $\beta = nL$ possible budget increments. One could equivalently normalize by the maximum budget $nL$, expressing performance as a function of the fraction of budget expended. This normalized representation is used in our plots, where the horizontal axis corresponds to the percentage of budget. One could alternatively normalize the horizontal axis by the number of unique individuals treated at each budget, yielding a per-patient notion of efficiency. This normalization is not effective because it is not aligned with the resource model in which costs increase per incremental level, so it would conflate different allocations that treat the same number of individuals but at different intensities.

Finally, ML-AUTOC has the same value interpretation as AUTOC for policies that induce a nested ranking over treatment actions, and can be extended as a cumulative performance metric for more general allocation policies. Naive policies such as \textsf{LFP}, \textsf{IFP}, and \textsf{PGP} as well as \textsf{PAG} produce an explicit ordering of $(i,\ell)$ actions, so the ML-AUTOC area can be interpreted as a cumulative comparison to the random precedence-feasible baseline as the budget expands. Although \textsf{PAO} does not generate a single nested ranking because it re-optimizes the allocation at each budget and may revise earlier decisions, its ML-AUTOC can still be computed as the cumulative average treatment effect relative to the random baseline across budgets. In this sense, ML-AUTOC provides a unified metric for comparing both nested and non-nested policies. In addition to ML-AUTOC, we also compare policies to \textsf{PAO} budget-wise by reporting the fraction of budgets for which one policy achieves a higher realized benefit than another (``win~\%''). This complementary metric highlights pointwise performance differences that may not be fully captured by cumulative area comparisons.

\subsection{Research Question}
\textit{Does a precedence-aware greedy policy achieve higher total average treatment effect (ATE) than naive ranking policies in multi-level treatment settings with precedence constraints?}

\noindent To investigate this, we examine the following empirical questions:
\begin{enumerate}
    \item \textbf{Policy Performance:} On synthetic datasets with different per-level treatment effect structures, how do \textsf{PAG} and \textsf{PAO} compare to naive policies in terms of realized total average treatment effect across budget levels?
    \item \textbf{Structural Behavior:} How do allocations produced by \textsf{PAO} which allows for re-allocations differ from those produced by nested greedy allocations \textsf{(PAG)}, and when does relaxing the monotonic allocation requirement lead to a higher realized average treatment effect?
\end{enumerate}

\section{Experiments}

\subsection{Datasets}

In order to evaluate ML-AUTOC under varying structural conditions, we generated 100 synthetic datasets of four types that were designed to isolate how different marginal benefit structures influence policy performance. Synthetic data was used for our experiment following \citet{kamran2024learning} since true individual-level treatment effects are rarely observable in practice. Since only one realized outcome was observed per individual, evaluating optimal treatment policies was infeasible using real data. Moreover, our framework assumes access to marginal treatment benefits, which further motivates the use of synthetic environments to simulate ground-truth performance under controlled assumptions. 

Each dataset contains $n$ individuals who may receive treatment levels $\ell \in \{1,2,\dots,L\}$ (with $L=4$ in all experiments) with the choice to treat a single individual to the next level $\ell$ being a unit of expenditure. Thus, any differences in model performance arise solely from the marginal benefit structure and from how rankings interact with precedence constraints. The datasets below were constructed to isolate how different marginal benefit structures interact with precedence constraints and to evaluate both the performance of ranking policies in terms of realized average treatment effect. In addition, their design facilitated the study of when naive, nested ranking heuristics fail and when precedence-aware, budget-adaptive policies can achieve higher realized benefit. In particular, they allowed us to:

\begin{itemize}
    \item RQ1: evaluate how the proposed \textsf{PAG} and \textsf{PAO} policies compare to naive policies in terms of realized average treatment effect across budgets,
    \item RQ2: evaluate when relaxing nested (prefix-based) ranking requirements, by allowing reallocation across budgets in \textsf{PAO}, leads to a higher realized total average treatment effect compared to the nested greedy \textsf{PAG} policy.
\end{itemize}

The full data-generating process, including distributional assumptions and implementation details for each dataset type, is provided in \cref{apd:data}.

\noindent \textbf{Strictly Increasing Marginal Benefit Dataset.}
In this dataset, higher treatment levels are always substantially more beneficial. To generate this, we first draw a baseline benefit for each level and each individual from a normal distribution. We then enforce monotonicity by increasing each successive level’s marginal benefit by at least one unit if the raw draw does not satisfy this. As a result, the marginal benefits always satisfy:
\(
\delta_i^{(\ell)} \;\ge\; \delta_i^{(\ell-1)} + 1
\). 

\noindent \textbf{Random Marginal Benefit Dataset.}
In this dataset, marginal benefits are nonnegative and have no ordering across treatment levels. For each individual $i$ and level $\ell$, the marginal benefit $\delta_i^{(\ell)}$ satisfies
\(\delta_i^{(\ell)} \ge 0\), 
but is otherwise unconstrained across levels. In particular, no monotonicity condition is imposed: higher treatment levels may yield greater, smaller, or comparable marginal benefit relative to earlier levels.
As a result, individuals may benefit most from early treatment intensities, later intensities, or intermediate levels, and these patterns can vary irregularly across both individuals and levels.

\noindent \textbf{Strictly Decreasing Marginal Benefit Dataset}.
In this dataset, marginal benefits are strictly decreasing across treatment levels, so earlier treatment intensities are always more beneficial than later ones. This dataset is constructed by taking the Strictly Increasing Marginal Benefit Dataset that appears as follows:
\[
\delta_i^{(1)} < \delta_i^{(2)} < \delta_i^{(3)} < \delta_i^{(4)}
\]
and applies a transformation so that:
\[
\tilde{\delta}_i^{(1)} = \delta_i^{(4)}, \quad
\tilde{\delta}_i^{(2)} = \delta_i^{(3)}, \quad
\tilde{\delta}_i^{(3)} = \delta_i^{(2)}, \quad
\tilde{\delta}_i^{(4)} = \delta_i^{(1)}.
\]
where $\tilde{\delta}_i^{(\ell)}$ denotes the marginal benefit at level $\ell$ in the strictly decreasing dataset.
Formally, the marginal benefits are constrained to satisfy:
\(
\delta_i^{(\ell)} \;\le\; \delta_i^{(\ell-1)} - 1
\)
for all individuals and levels.

\noindent \textbf{Heterogeneous Strictly Increasing Marginal Benefit Dataset.}
In this dataset, marginal benefits remain strictly increasing across treatment levels for each individual, but the scale of these benefits varies across individuals. Unlike the homogeneous strictly increasing dataset, individuals are not drawn from a common marginal benefit distribution. Instead, each individual $i$ has a scale parameter $s_i$ drawn from a log-normal distribution, and marginal benefits are generated sequentially across levels subject to $\delta_i^{(\ell)} \ge \delta_i^{(\ell-1)} + 1$. Because individuals differ in scale, lower treatment levels of one individual may exceed higher treatment levels of another (e.g., $\delta_i^{(2)} > \delta_j^{(3)}$), so within-individual monotonicity is preserved while the relative ranking of marginal increments across individuals and levels is nontrivial. This setting is useful for studying the trade-off between completing treatment for high-potential individuals and allocating early treatment levels across many individuals.


\subsection{RQ1: Strictly Increasing Marginal Benefit Dataset}

\begin{figure}[ht]
    \centering
    \includegraphics[width=1\linewidth]{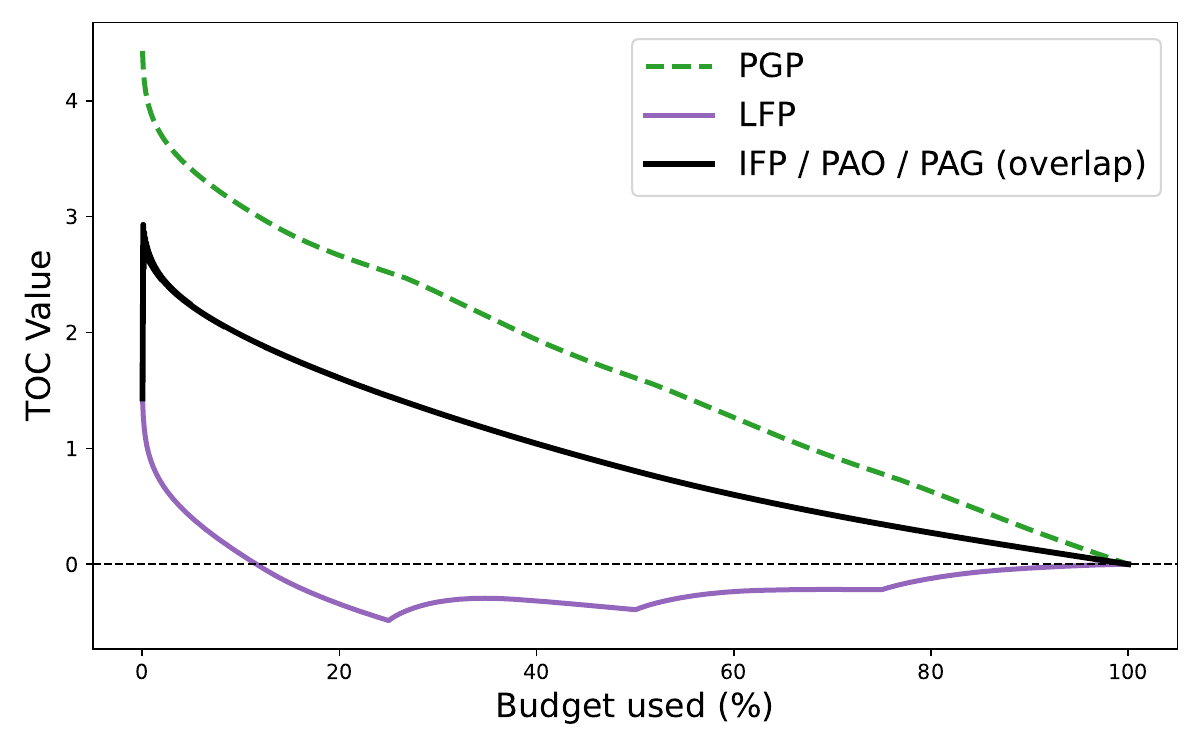}
    \vspace{-1em}
    \caption{Average TOC curves across budgets for the strictly increasing marginal benefit dataset. Curves are averaged over 100 datasets and plotted relative to a random baseline. $\pm$ one standard deviation error bars are plotted but too small to be visually distinguishable.}
    \label{fig:2}
    \vspace{-1em}
\end{figure}

As a result of the structure of the strictly increasing dataset, policies that prioritize allocating higher levels early achieve a substantially higher average treatment effect across budgets. Among policies that respect the precedence structure (i.e., all policies except \textsf{PGP}), both \textsf{IFP} and \textsf{PAG} achieve similar performance. 
To further illustrate how policies behave across the full budget range, we examine the average ML-TOC curves aggregated across all datasets. \cref{fig:2} plots the mean targeting operating characteristic for each policy, with budgets on the horizontal axis and average treatment effect relative to a random baseline on the vertical axis.
Consistent with the aggregate results in \cref{tab: RQ1-ML-AUTOC-strict}, the average curves reveal that \textsf{IFP}, \textsf{PAO}, and \textsf{PAG} exhibit nearly identical performance across all budget levels.

\begin{table}[ht]
    \centering
    \caption{ML-AUTOC Values for Strictly Increasing Marginals}
    \label{tab: RQ1-ML-AUTOC-strict}
    \footnotesize
    \setlength{\tabcolsep}{4pt}
    \begin{tabular}{lc}
    \toprule
    Policy & ML-AUTOC (mean $\pm$ std)\\
    \midrule
    PGP & 1.666352 $\pm$ 0.003744 \\
    IFP & 0.941827 $\pm$ 0.009646 \\
    LFP & -0.154488 $\pm$ 0.003209 \\
    PAG & 0.940726 $\pm$ 0.009646 \\
    PAO & 0.941828 $\pm$ 0.009646 \\
    \bottomrule
    \end{tabular}
\end{table}

The \textsf{LFP} curve remains distinctly separated from the other policies, not only due to its lower overall performance, but also because of its characteristic shape. This reflects the fact that \textsf{LFP} fully allocates lower treatment levels across individuals before advancing to higher levels and thus spends early budgets on the lowest-level, smallest marginal benefits. 

Overall, in settings with strictly ordered marginal benefits, these findings indicate that allowing flexible, non-nested allocation (as in \textsf{PAO}) provides no additional benefit in strictly increasing settings, where the optimal allocation is inherently nested and can be naively achieved by \textsf{IFP}.

\subsection{RQ1: Random Marginal Benefit Dataset}

Unlike in the strictly increasing case, in the random marginal case clear separation emerges between policies as the budget increases. The flexible precedence-aware policy \textsf{PAO} consistently dominates all other valid policies. 

\begin{figure}[ht]
    \centering
    \includegraphics[width=1\linewidth]{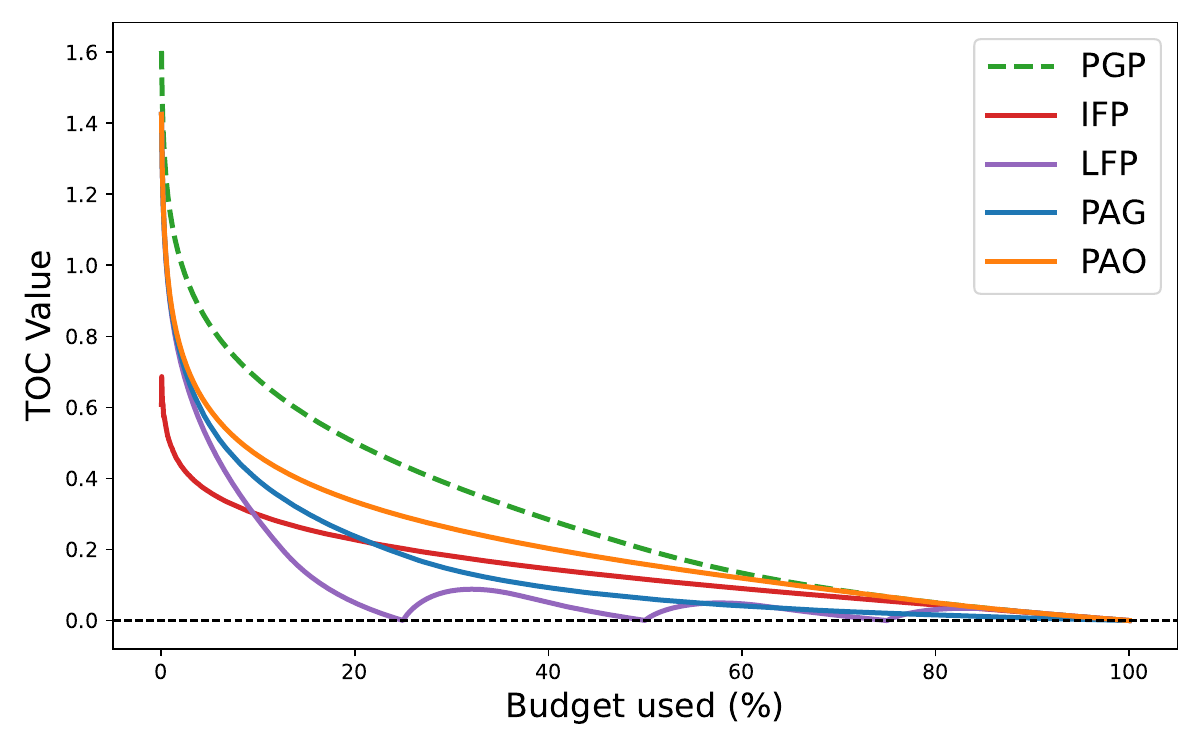}
    \vspace{-1em}
    \caption{Average TOC curves across budgets for the random marginal benefit dataset. Curves are averaged over 100 datasets and plotted relative to a random baseline. $\pm$ one standard deviation error bars are plotted but too small to be visually distinguishable.}
    \label{fig:3}
    \vspace{-2em}
\end{figure}

While the greedy policy \textsf{PAG} initially outperforms the global score policy \textsf{IFP}, the two curves intersect at approximately 25\% of the total budget ($\beta \approx 1000/4000$). Beyond this point, \textsf{IFP} achieves higher realized average treatment effect for the remainder of the budget range. This crossover highlights the cost of enforcing nested (prefix-based) rankings when marginal benefits vary across treatment levels, as early allocation decisions restrict the ability of \textsf{PAG} to adapt at later budgets. This suggests that in the case of smaller budgets and random marginal benefits, PAG may be a better policy to pursue to allocate resources.

In contrast, \textsf{PAO} avoids this tradeoff entirely by allowing non-nested, dynamically optimized allocations, enabling it to select the highest marginal benefit action at each budget. As a result, \textsf{PAO} maintains superior performance across nearly all budget levels. 

Between precedence-respecting policies, \textsf{PAG} and \textsf{IFP} achieve nearly identical ML-AUTOC values, despite exhibiting different behavior across the budget range in \cref{fig:3}. This highlights an important distinction between pointwise performance and cumulative performance: although \textsf{IFP} overtakes \textsf{PAG} at larger budgets, the early-budget advantage of \textsf{PAG} is sufficient to equalize total realized benefit when aggregated across all budgets. However, \textsf{PAO} achieves a substantially higher ML-AUTOC than both \textsf{PAG} and \textsf{IFP}, indicating that flexible allocation improves not only pointwise performance but also cumulative performance when marginal benefits are heterogeneous.

\begin{table}[ht]
    \centering
    \caption{ML-AUTOC Values for Random Marginals}
    \footnotesize
    \setlength{\tabcolsep}{4pt}
    \begin{tabular}{lc}
    \toprule
    Policy & ML-AUTOC (mean $\pm$ std)\\
    \midrule
    PGP & 0.285943 $\pm$ 0.005926 \\
    IFP & 0.141575 $\pm$ 0.004947 \\
    LFP & 0.096445 $\pm$ 0.003341 \\
    PAG & 0.141024 $\pm$ 0.004727 \\
    PAO & 0.209599 $\pm$ 0.004183 \\
    \bottomrule
    \end{tabular}
    \label{tab:RQ1 ML-AUTOC2}
\end{table}
Overall, in the random marginal benefit setting, relaxing nested ranking constraints enables higher pointwise performance across budgets, while cumulative ML-AUTOC comparisons reveal tradeoffs between early- and late-budget performance among nested policies.

\subsection{RQ1: Strictly Decreasing Marginal 
Benefit Dataset}

\begin{table}[ht]
    \centering
    \caption{ML-AUTOC Values for Strictly Decreasing Marginals}
    \footnotesize
    \setlength{\tabcolsep}{4pt}
    \begin{tabular}{lc}
    \toprule
    Policy & ML-AUTOC (mean $\pm$ std)\\
    \midrule
    PGP & 0.350069 $\pm$ 0.003820 \\
    IFP & -0.368648 $\pm$ 0.009757 \\
    LFP & 0.348233 $\pm$ 0.003348 \\
    PAG & 0.350069 $\pm$ 0.003820 \\
    PAO & 0.350069 $\pm$ 0.003820 \\
    \bottomrule
    \end{tabular}
    \label{tab:RQ1 ML-AUTOC3}
\end{table}

In the strictly decreasing marginal benefit setting, \textsf{PGP}, \textsf{PAG}, \textsf{PAO}, and \textsf{LFP} exhibit nearly identical performance. This occurs because optimal allocations in this setting prioritize exhausting high early marginal benefits, leaving little room for gains from reallocating treatment across individuals at later budgets. \textsf{IFP} performs worse because it is forced to treat an entire individual and therefore is unable to capitalize on larger marginals earlier. As a result, flexibility provides no meaningful advantage when marginal benefits are strictly front-loaded. An average plot, similar to the one created for the strictly increasing and random marginal dataset is included in \cref{apd:SD}.

\subsection{RQ1: Heterogeneous Strictly Increasing Marginal Benefit Dataset}

The heterogeneous strictly increasing marginal benefit setting introduces substantial overlap in the ranking of marginal treatment increments by \textsf{PAO}, \textsf{PAG}, and \textsf{IFP}. The ML-AUTOC values in \cref{tab:RQ1-ML-AUTOC-het} show that \textsf{IFP}, \textsf{PAO}, and \textsf{PAG} achieve nearly identical performance. The differences across these policies are negligible both in mean performance and variance, indicating that the additional flexibility of \textsf{PAO} does not translate into meaningful gains.

\begin{table}[ht]
    \centering
    \caption{ML-AUTOC Values for Heterogeneous Strictly Increasing Marginals}
    \label{tab:RQ1-ML-AUTOC-het}
    \footnotesize
    \setlength{\tabcolsep}{4pt}
    \begin{tabular}{lc}
    \toprule
    Policy & ML-AUTOC (mean $\pm$ std)\\
    \midrule
    PGP & 1.901681 $\pm$ 0.019583 \\
    IFP & 1.394540 $\pm$ 0.028555 \\
    LFP & 0.003476 $\pm$ 0.008919 \\
    PAG & 1.392382 $\pm$ 0.028573 \\
    PAO & 1.394548 $\pm$ 0.028554 \\
    \bottomrule
    \end{tabular}
\end{table}

\begin{figure}[ht]
    \centering
    \includegraphics[width=1\linewidth]{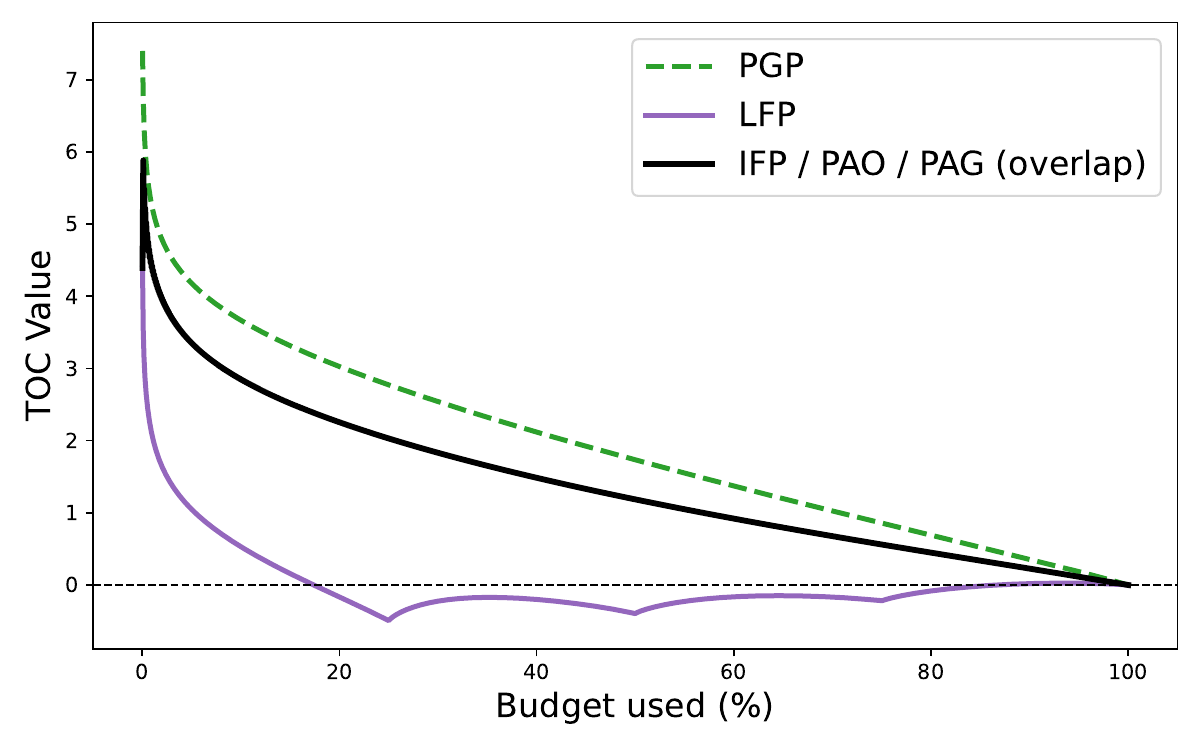}
    \vspace{-1em}
    \caption{Average TOC curves across budgets for the heterogeneous strictly increasing marginal benefit dataset. Curves are averaged over 100 datasets and plotted relative to a random baseline. $\pm$ one standard deviation error bars are plotted but too small to be visually distinguishable.}
    \label{fig:het}
    \vspace{-1em}
\end{figure}

Consistent with \cref{tab:RQ1-ML-AUTOC-het}, the curves for \textsf{IFP}, \textsf{PAO}, and \textsf{PAG} overlap and are visually indistinguishable across all budget levels (\cref{fig:het}). This overlap indicates that, despite cross-individual heterogeneity, all three policies induce nearly identical rankings of marginal treatment increments. While heterogeneity introduces crossings in the global ordering, it does not fundamentally disrupt the prefix structure of the optimal allocation. As a result, policies that enforce nested allocations remain sufficient to capture the majority of attainable gains.

\subsection{RQ2}

\cref{tab:RQ2-structural} quantifies the structural differences between flexible and nested precedence-aware policies by measuring how often \textsf{PAO} strictly outperforms competing policies across budgets.

\begin{table}[ht]
    \centering
    \caption{Frequency with which \textsf{PAO} strictly outperforms baseline policies across budgets.}
    \label{tab:RQ2-structural}
    \footnotesize
    \setlength{\tabcolsep}{6pt}
    \scalebox{0.72}{
    \begin{tabular}{lcc}
        \toprule
        Dataset & $\%\,$Budgets PAO $>$ IFP & $\%\,$Budgets PAO $>$ PAG \\
        \midrule
        Strictly Increasing 
        & $45.41 \pm 31.32$ 
        & $87.65 \pm 7.82$ \\
        Random 
        & $96.70 \pm 3.15$ 
        & $99.55 \pm 0.26$ \\
        Strictly Decreasing
        & $99.94 \pm 0.02$ 
        & $44.44 \pm 29.59$ \\
        Heterogeneous Increasing
        & $47.92 \pm 26.15$
        & $94.17 \pm 3.83$ \\
        \bottomrule
    \end{tabular}}
\end{table}

In settings with strictly increasing marginal benefits, allowing reallocation across budgets yields frequent improvements over \textsf{PAG}, but does not translate into consistent gains over \textsf{IFP}. In contrast, when marginal benefits are random and heterogeneous across levels, relaxing the nested (prefix-based) constraint enables \textsf{PAO} to consistently dominate both \textsf{IFP} and \textsf{PAG} across nearly all budget levels. In this regime, the ability to reallocate across individuals and levels is critical, as the globally optimal sequence of marginal increments does not follow a simple prefix structure. When marginal benefits are strictly decreasing, \textsf{PAO} almost always outperforms \textsf{IFP} but offers no systematic advantage over \textsf{PAG}, reflecting the fact that optimal allocations in this setting favor concentrating resources early rather than flexibly reallocating across individuals. In this case, the optimal policy is naturally nested, so flexibility provides little additional benefit. Finally, in the heterogeneous strictly increasing setting, the conclusions are similar to the strictly increasing setting: allowing reallocation leads to consistent improvements over \textsf{PAG} at individual budgets but not over \textsf{IFP}.

\section{Discussion and Conclusion}

\subsection{PAO Versus PAG}

As mentioned in \cref{section 2.6}, \textsf{PAO} is uniquely different from the other policies in that it does not have a direct interpretable ranking that can be derived from the policy allocation. Except in the strictly decreasing setting, \textsf{PAO} outperforms \textsf{PAG} at the majority of budget levels, with particularly strong dominance in the random marginal benefit setting. This improvement comes at the cost of interpretability. Because allocations produced by \textsf{PAO} are not nested, following the policy at budget $\beta$ does not guarantee that the treatment chosen at budget $\beta + 1$ extends the previous allocation in a greedy manner. In particular, maximizing realized average treatment effect may require reallocating budget away from partially treated individuals toward initiating treatment for others, breaking the prefix structure that underlies ranking-based approaches such as \textsf{PAG}.

When individuals can receive multiple treatment intensities, optimal allocation may require switching between individuals and treatment levels as the budget evolves, rather than committing to a fixed ordering. \textsf{PAO} captures this flexibility, demonstrating that in multi-level intervention problems, maximizing realized treatment effect can necessitate departing from nested ranking structures that are sufficient in the binary AUTOC setting.

\subsection{Time Analysis}

\begin{figure}[ht]
    \centering
    \caption{Empirical runtime scaling of ranking policies as the number of individuals $n$ increases, measured on strictly increasing marginal benefit datasets with $L=4$.}
    \includegraphics[width=\linewidth]{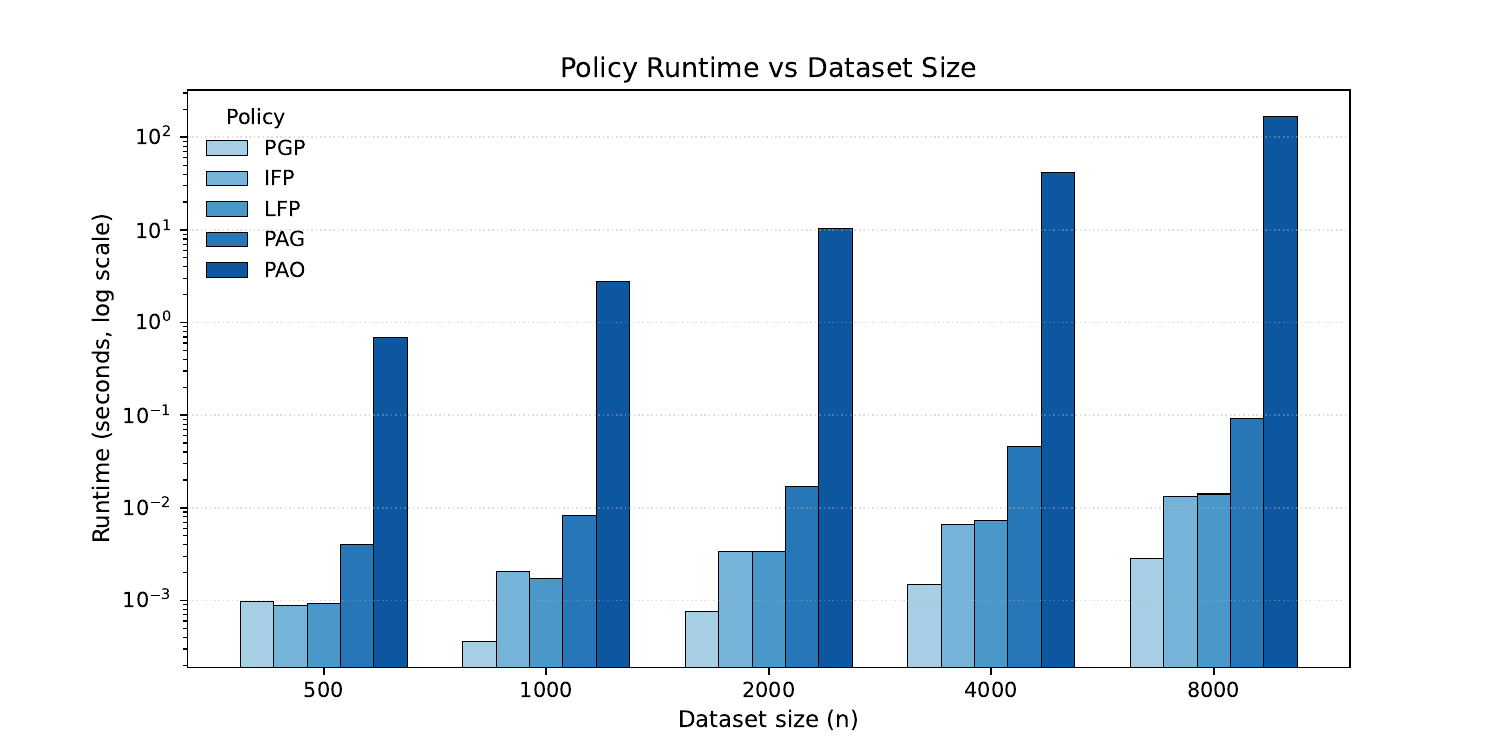}
    \label{fig:timing}
\end{figure}

To evaluate the computational cost of each ranking, we measure the runtime of each ranking on a strictly increasing marginal benefit dataset of depth $L=4$ and varying population size \textit{n}. \cref{fig:timing} and \cref{tab:timing} report the time required to generate a full ranking for each policy at each size \textit{n}.

The naive ranking policies of \textsf{LFP}, \textsf{IFP}, and \textsf{PGP} exhibit negligible runtimes across the dataset sizes considered. While these policies differ in how rankings are constructed, each reduces the multi-level allocation problem to a small number of comparison-based sorting operations, with the number of sorts bounded above by the treatment depth $L$. As a result, their computational cost is dominated by standard sorting steps whose total complexity scales near-linearly in the number of individuals. For fixed treatment depth $L$, this leads to effective runtimes that grow approximately on the order of $\mathcal{O}(n \log n)$, consistent with the empirical timing results shown in \cref{fig:timing}. A detailed breakdown of the time complexity for each policy is provided in \cref{apd:bigoh}.

Between \textsf{PAG} and \textsf{PAO}, \textsf{PAG} consistently achieves a lower runtime. This difference is substantial, with \textsf{PAG} running orders of magnitude faster than \textsf{PAO} in practice. This gap arises from the dynamic programming procedure used by \textsf{PAO} to compute a globally optimal allocation independently at each budget level. While our results show that this flexibility can improve solution quality, it introduces \textit{substantial} computational overhead, limiting its usefulness. 

These results highlight a clear tradeoff between computational efficiency and allocation flexibility. While \textsf{PAO} provides the most reactive policy by re-allocating every budget level, its runtime makes it impractical for large-scale cases. In contrast, \textsf{PAG} scales comparably to the naive ranking policies while still enforcing precedence constraints. As a result, \textsf{PAG} likely is the preferred policy in applications where computational constraints are binding, and rankings must be generated at scale.

\subsection{When Does Flexibility Matter?}

Our experiments suggest that the value of flexibility in allocation relies on the structure of marginal benefits. Specifically, to what extent the optimal decision at multiple budget levels $\beta$ in a cohort requires switching treatments.

In the strictly increasing marginal benefit setting (heterogeneous and homogeneous), treating higher levels as early as possible is the most beneficial to overall score. This intuition is verified by our empirical observation that several policies exhibited the same behavior: treating an individual fully once started. This behavior reduced the importance of reallocation as treating a higher level provided more benefit than reallocating. Consequently, \textsf{PAG} and \textsf{PAO} do not achieve a meaningful improvement over strong naive baselines such as \textsf{IFP}.

In the random marginal benefit setting, early commitments to treating marginals can become costly. For example, treating an individual to an intermediate level may be dominated by initiating a full treatment on a different individual during a subsequent budget level. This benefit structure is precisely the regime where the nested (prefix-based) constraint becomes binding. \textsf{PAG} can perform well at early budgets by selecting the best feasible next step greedily, but its nested allocation requirement limits its ability to revise earlier choices as the budget grows. By recomputing the allocation at each budget, \textsf{PAO} can reallocate away from partially treated individuals when a better frontier emerges, which explains its consistent pointwise dominance across budgets.

In the strictly decreasing marginal benefit setting, most of the valuable treatment marginals are able to be treated first, and therefore the best corresponding policy exhausts earlier treatment levels across many individuals. Because later levels are consistently less valuable than earlier levels, there is little benefit to reallocation. This explains why \textsf{PAO} offers no systematic advantage over \textsf{PAG} in this regime, even though it can still outperform \textsf{IFP} due to \textsf{IFP}'s commitment to fully treating individuals.

Overall, these findings clarify when multi-level allocation meaningfully departs from the binary treatment setting. In binary allocation, policies naturally correspond to their ranking output, so a policy to allocate treatment effects further, such as a policy in our case, is unnecessary. In the multi-level setting, this equivalence can break down. When marginal benefits vary across treatment levels, maximizing realized treatment effect may require policies that change which individuals they prioritize as the budget evolves. In such cases, allocation cannot be represented by a single nested ranking, and flexible policies such as \textsf{PAO} provide a structural advantage.

\subsection{Non-Nested Allocations In The Clinical Setting}

A potential limitation of non-nested allocation policies such as \textsf{PAO} is their lack of interpretability and feasibility in real-world clinical settings. For example, in a setting where a hospital has already allocated treatment up to its current budget, the arrival of an additional unit of budget introduces ambiguity in how that resource should be deployed. Unlike nested policies, which extend naturally by following a fixed ranking, \textsf{PAO} may require re-solving the allocation problem entirely, potentially revising earlier decisions. This raises a real question of whether an incremental increase should trigger a reallocation or whether they should remain stable once allocated.

In practice, we propose that \textsf{PAO} should not be implemented as a full dynamic reallocation at each budget increment. Rather, allocation decisions can be made at regular intervals, such as shift changes in hospitals, when resources are known or fixed. This avoids the need to re-optimize allocations in response to small, incremental changes in budget, where solving a full allocation problem to assign a single additional treatment may be impractical.

We encourage this problem to be explored further in future work, particularly in designing policies that balance allocation flexibility with the need for stability and interpretability in clinical environments.

\subsection{On The Use Of Non-Synthetic Data}

A natural question that arises at the end of this research is whether the findings in this paper can be extended to real-world data. While evaluating allocation policies on observational data is highly desirable, the implementation of the multi-level treatment setting on non-synthetic data is difficult.

First, the evaluation of allocation policies in our framework requires access to individual-level marginal treatment effects for each treatment level. In real-world datasets, these quantities are inherently unobserved. Estimating these quantities requires strong assumptions of consistency, unconfoundedness, and overlap across all treatment levels \citep{Rubin1974}. In multi-level settings, the overlap assumption becomes particularly restrictive, as each individual must have a non-zero probability of receiving every treatment intensity.

Even when these assumptions hold, estimating marginal treatment effects at multiple levels introduces statistical noise. Because the multi-level setting requires estimating a sequence of incremental effects, the errors in the compounding of these estimates make it difficult to disentangle where policy performance comes from.

That said, extending this work to non-synthetic data is important future work. One promising approach could be the use of semi-synthetic data where real covariates and treatment assignments are combined with simulated outcome models. More broadly, applying the ML-AUTOC framework with high-quality causal estimators and explicitly accounting for estimation uncertainty will be critical for understanding how these allocation policies perform in practice.

\subsection{Conclusion}

This paper investigates whether precedence-aware greedy and reallocation policies outperform naive ranking-based policies in terms of achieving higher realized total average treatment effect in multi-level treatment settings. Our findings show that reallocation for optimal benefit (\textsf{PAO}) is not necessary to achieve strong performance. In settings with strictly ordered marginal benefits, \textsf{PAG} achieves performance comparable to naive ranking-based policies.

However, in settings where there are random marginal benefits for different treatment levels, the relaxation of the nesting constraint for \textsf{PAO} is shown to have higher pointwise performance across different budgets, indicating that in these settings, it is important to maximize the realized treatment effect by reallocating resources rather than relying on a naive approach.

\section*{Acknowledgments}
This work was completed as part of the Honors Thesis of Olin Gilster (\url{https://etd.library.emory.edu/concern/etds/5138jg33v}). The authors would like to thank Ruoxuan Xiong and Michelangelo Grigni for their helpful feedback.

\section*{AI Use Statement}
We used generative AI tools to assist with copy-editing of the final manuscript. The authors take responsibility for the final content of this work, including text, claims, or artifacts produced with the aid of generative AI. 

\bibliography{ref}

\clearpage

\appendix
\crefalias{section}{appendix}
\crefalias{subsection}{appendix}
\onecolumn

\section{}\label{apd:first}

\subsection{Heterogeneous Treatment Costs}\label{apd:costs}

The primary analysis in this paper assumes that each marginal treatment level incurs a unit cost. While this simplifies the allocation problem and isolates the role of marginal benefit structure, it abstracts away an important feature of many real-world intervention systems: treatment intensity often varies in cost.

In practice, higher levels of intervention typically require greater resource expenditure. For example, escalating clinical treatment intensity may require additional staff time, specialized services, or longer duration of care \citep{Drummond2015}. Cost heterogeneity is therefore not merely a modeling convenience but reflects structural properties of real-world resource allocation problems. Allowing treatment levels to differ in cost aligns the framework more closely with standard welfare maximization and cost-effectiveness formulations in economics and operations research, where decision-makers face a fixed budget and must allocate heterogeneous interventions to maximize total benefit \citep{BertsimasTsitsiklis1997}.

To start, let $c_\ell > 0$ denote the marginal cost of advancing an individual from treatment level $\ell-1$ to level $\ell$. We define the cumulative cost of assigning individual $i$ to level $\ell$ as
\[
C(\ell) = \sum_{k=1}^{\ell} c_k,
\]
which is weakly increasing in $\ell$. Given marginal benefits $\delta_i^{(\ell)}$ as defined in \cref{eq:1}, the allocation problem under a fixed total budget $B$ becomes:
\[
\max_{\{z_{i,\ell}\}} \sum_{i=1}^{n}\sum_{\ell=1}^{L} \delta_i^{(\ell)} z_{i,\ell}
\]
subject to
\[
\sum_{i=1}^{n}\sum_{\ell=1}^{L} c_\ell z_{i,\ell} \leq B, \qquad
z_{i,\ell} \leq z_{i,\ell-1} \;\; \forall i,\, \ell \geq 2, \qquad
z_{i,\ell} \in \{0,1\}.
\]
The second constraint enforces precedence: an individual may only receive level $\ell$ if all prior levels have been assigned. The problem now resembles a precedence-constrained knapsack problem, where each marginal treatment increment yields benefit $\delta_i^{(\ell)}$ but consumes $c_\ell$ units of budget \citep{Pisinger2005}. The individual optimizing must therefore trade off both benefit magnitude and cost-effectiveness.

Under unit costs, the budget parameter $\beta$ directly corresponds to the number of marginal treatment increments allocated. Each step in the ML-TOC curve therefore reflects one additional unit of treatment. With heterogeneous costs, this interpretation changes. The budget $B$ now measures total expenditure rather than number of treatment increments. A single additional allocation may consume multiple units of budget, and different increments may advance the budget frontier by unequal amounts. Consequently:
\begin{enumerate}
\setlength{\topsep}{0pt}
\setlength{\itemsep}{0pt}
    \item The horizontal axis of ML-TOC curves would represent cumulative expenditure rather than count of treatment steps.
    \item Budget increments become non-uniform.
    \item The mapping between allocation rank and budget becomes nonlinear.
\end{enumerate}

We believe that introducing heterogeneous costs should increase the importance of flexible, non-nested allocation policies. Under unit costs, the primary trade-off concerns only marginal benefit magnitude. However, when costs differ, optimal allocation depends on benefit-to-cost ratios $\delta_i^{(\ell)} / c_\ell$. If higher treatment levels have disproportionately high costs, it may become optimal to partially treat multiple individuals rather than fully treat a small number. Conversely, if early levels are inexpensive but later levels are costly, the optimal policy may exhibit switching behavior across individuals to maximize aggregate benefit per dollar spent. We therefore hypothesize:

\begin{quote}
When treatment costs are heterogeneous, precedence-aware policies that allow reallocation (e.g., \textsf{PAO}) will outperform ranking-based or nested greedy policies by a larger margin than in the unit-cost setting.
\end{quote}
In particular, heterogeneous costs introduce an additional dimension along which early allocation decisions may become suboptimal at later budget levels, increasing the value of dynamic reoptimization.

\subsection{Naive Policy Allocation Matrices}

Naive policies in the multi-level treatment framework benefit from collapsing the multi-level treatment structure into a one-time ordering problem. 

We illustrate this flattening using the toy marginal benefit matrix in \cref{tab:toyexample2}.

\begin{table}[ht]
    \centering
    \caption{True marginal benefit matrix (see also \cref{tab:1})}
    \label{tab:toyexample2}
    \begin{tabular}{|c|c|c|} \hline
    $\delta_i^{(\ell)}$ & $\ell = 1$ & $\ell = 2$ \\ \hline
    $x_1$ & 1 & 5 \\ \hline
    $x_2$ & 2 & 3 \\ \hline
    \end{tabular}
\end{table}

Each policy maps this $2 \times 2$ matrix into a length-4 ranking over treatment actions $(i,\ell)$, effectively flattening the multi-level structure.

\begin{enumerate}

\item \textsf{LFP}\\
\textsf{LFP} sorts individuals independently within each treatment level and then concatenates the level-wise rankings.  
For $\ell = 1$, we have $x_2 \succ x_1$ since $2 > 1$; for $\ell = 2$, we have $x_1 \succ x_2$ since $5 > 3$.  
The resulting flattened allocation matrix is:
\[
\begin{bmatrix}
(x_2,1)\\
(x_1,1)\\
(x_1,2)\\
(x_2,2)
\end{bmatrix}
\quad \text{with values} \quad
[2,\;1,\;5,\;3].
\]

\item \textsf{IFP}\\
\textsf{IFP} first aggregates marginal benefits across levels for each individual,
\[
\sum x_1 = 1 + 5 = 6, \qquad \sum x_2 = 2 + 3 = 5,
\]
and ranks individuals by this total score. Treatment actions are then expanded in precedence order for each individual.  
The resulting flattened allocation matrix is:
\[
\begin{bmatrix}
(x_1,1)\\
(x_1,2)\\
(x_2,1)\\
(x_2,2)
\end{bmatrix}
\quad \text{with values} \quad
[1,\;5,\;2,\;3].
\]

\item \textsf{PGP}\\  
\textsf{PGP} ignores precedence at ranking time and globally sorts all $(i,\ell)$ pairs by their marginal benefit values.  
Sorting $\{1,5,2,3\}$ in descending order yields:
\[
\begin{bmatrix}
(x_1,2)\\
(x_2,2)\\
(x_2,1)\\
(x_1,1)
\end{bmatrix}
\quad \text{with values} \quad
[5,\;3,\;2,\;1].
\]

\end{enumerate}

\subsection{LFP Code}\label{apd:LFP}
\begin{verbatim}
# Level-First Policy (LFP)

initialize empty list ranking

for each level l from 1 to L:

    # sort individuals by marginal benefit at this level
    create list of individuals sorted by delta[i][l] in descending order

    for each individual i in this sorted list:
        append (i, l) to ranking

return ranking
\end{verbatim}
\subsection{IFP Code}\label{apd:IFP}
\begin{verbatim}
# Individual-First Policy (IFP)

initialize empty list ranking

# compute total value for each individual
for each individual i:
    total[i] = sum over l of delta[i][l]

# sort individuals by total value
create list of individuals sorted by total[i] in descending order

# assign all levels for each individual before moving on
for each individual i in sorted list:
    for each level l from 1 to L:
        append (i, l) to ranking

return ranking
\end{verbatim}
\subsection{PGP Code}\label{apd:PGP}
\begin{verbatim}
# Pooled Global Policy (PGP)

initialize empty list of pairs

# treat every (i, l) as a separate item
for each individual i:
    for each level l:
        add (i, l) to list

# sort all pairs by marginal benefit
sort list of pairs by delta[i][l] in descending order

ranking = sorted list
return ranking
\end{verbatim}

\subsection{PAG Code}\label{apd:PAG}
\begin{verbatim}
# Precedence-Aware Greedy (PAG)

initialize empty ranking list

# priority queue stores (i, l) with priority delta[i][l]
initialize empty max-priority queue

# initially, only level 1 is feasible for each individual
for each individual i:
    push (i, 1) into priority queue with priority delta[i][1]

# iterate over full budget
for b = 1 to nL:
    # select best feasible marginal
    (i, l) = pop max from priority queue

    add (i, l) to ranking

    # unlock next level for that individual
    if l + 1 <= L:
        push (i, l+1) into priority queue
        with priority delta[i][l+1]

return ranking
\end{verbatim}
\subsection{Big-Oh Analysis}\label{apd:bigoh}

We analyze the computational complexity of the naive ranking policies as a function of the number of individuals $n$ and treatment levels $L$. Our budget $\beta$ represents our total treatment actions. 

Our analysis relies on well-known results for comparison-based sorting that establish a worst-case lower bound of $\Omega(k \log k)$ comparisons for sorting $k$ keys, with matching $O(k \log k)$ upper bounds achievable by standard algorithms \citep{Knuth1973}.

\begin{enumerate}
\item \textsf{LFP} \\
The implementation of \textsf{LFP} in \cref{apd:LFP} consists of an outer loop over levels $\ell=1,\dots,L$.\\
The policy computes a sort for each level which costs $O(n\log n)$ under comparison-based sorting. The subsequent inner loop iterates once over the sorted indices and writes $n$ entries into the ranking array, costing $O(n)$. Summing over all levels yields $O(L\,n\log n)$.
Since the array writes are asymptotically dominated by the sorting step, the total running time of \textsf{LFP} is:
\[
O(n L\log n)
\]

\item \textsf{IFP} \\
The implementation of \textsf{IFP} in \cref{apd:IFP} first computes an aggregate score for each individual by summing marginal benefits across all levels. This requires iterating over the full $(n \times L)$ matrix once, costing $O(nL)$. \\
The policy then sorts the $n$ aggregate scores, which costs $O(n\log n)$ under comparison-based sorting. Finally, the nested loop expands each individual into its $L$ treatment actions and writes $nL$ entries into the ranking array, costing $O(nL)$. Combining terms yields a total running time of
\[
O(nL + n\log n).
\]

\item \textsf{PGP} \\ 
The implementation of \textsf{PGP} in \cref{apd:PGP} first flattens the $(n \times L)$ marginal benefit matrix into a length $nL$ vector, which costs $O(nL)$. \\
The policy then sorts these marginal benefit values, which costs $O(nL\log(nL))$ under comparison-based sorting. Constructing the ranking array from the sorted indices requires writing $nL$ entries and therefore costs $O(nL)$. The sorting step dominates, so the total running time of \textsf{PGP} is $O(nL\log(nL))$.

\item \textsf{PAG} \\
The implementation of \textsf{PAG} in \cref{apd:PAG} selects treatment actions sequentially over the full budget horizon of $nL$ marginal increments. Rather than scanning all $n$ individuals at each step, the algorithm maintains the current feasible frontier in a priority queue. Initially, the queue contains the first feasible marginal benefit for each of the $n$ individuals. \\
At each budget step, the algorithm removes the maximum feasible marginal benefit from the priority queue, which costs $O(\log n)$. If the selected action corresponds to level $\ell$ for individual $i$, then level $\ell+1$ for that same individual becomes newly feasible and is inserted into the priority queue, which also costs $O(\log n)$. Thus, each budget step requires at most one extract-max and one insert operation, for a total cost of $O(\log n)$ per step. Since there are $nL$ total treatment actions, and initializing the priority queue costs $O(n)$ (dominated by the main loop), the total running time is
\[
O(nL \log n).
\]

\item \textsf{PAO} \\
The implementation of \textsf{PAO} first computes prefix sums across treatment levels for each individual, which costs $O(nL)$. The policy then applies a dynamic program over individuals and budgets $nL$, updating an array of length $nL$ for each of $L$ possible allocation sizes per individual. This yields a total dynamic programming cost of $O(nL nL)$. Therefore, the overall running time of \textsf{PAO} is
\[
O(n^2 L^2).
\]

\end{enumerate}

\subsection{Big-Oh Table}

\begin{table}[ht]
\centering
\caption{Asymptotic runtime of ranking and precedence-aware policies as a function of the number of individuals $n$ and treatment levels $L$.}
\begin{tabular}{|l|c|}
\hline
\textbf{Policy} & \textbf{Runtime (Big-Oh)} \\
\hline
\textsf{PGP}    & $O (nL \log(nL))$ \\
\textsf{IFP}    & $O(nL + n\log n)$ \\
\textsf{LFP}   & $O(nL\log n)$ \\
\textsf{PAG} &  $O(nL \log n)$ \\
\textsf{PAO}  &  $O(n^2 L^2)$ \\
\hline
\end{tabular}
\end{table}

\subsection{Dynamic Programming}\label{apd:dynamic}
Dynamic programming was used to implement the \textsf{PAO} policy. The formulation is closely related to classical knapsack-style dynamic programming, in which a finite resource budget is allocated across multiple items to maximize total value \citep{Bellman:DynamicProgramming, KellererPferschyPisinger2004}. \\
In this setting, each individual corresponds to a group of mutually exclusive choices indexed by $t_i \in \{0,1,\dots,L\}$, and the total budget is $nL$ \citep{KellererPferschyPisinger2004}. Dynamic programming provides an exact solution by tracking the optimal achievable benefit for each intermediate budget level. \\
In order to use this approach, prefix sums of marginal benefits are computed, which allows for a constant-time computation of the assignment of treatment levels to a particular individual. The dynamic program iterates through individuals and prefix lengths, updating the value function on all budgets. The transition of the value function is shown in the following code:
\begin{verbatim}
# PAO: Dynamic programming over individuals and budgets

n = number of individuals
L = number of treatment levels
B = n * L

# Step 1: compute prefix values
# prefix[i][m] = total value of assigning the first m levels to individual i
initialize prefix as an n x (L + 1) table of zeros

for each individual i:
    prefix[i][0] = 0
    for each level m from 1 to L:
        prefix[i][m] = prefix[i][m - 1] + delta[i][m]

# Step 2: initialize dynamic program
# dp_prev[b] = best total value achievable using processed
# individuals and budget b
initialize dp_prev[0] = 0
initialize dp_prev[b] = -infinity for all b > 0

# choice[i][b] will store how many levels were assigned to individual i
# in the optimal solution at budget b during the i-th step
initialize choice as an n x (B + 1) table of zeros

# Step 3: process individuals one at a time
for each individual i:

    # start with the option of assigning no treatment to this individual
    best[b] = dp_prev[b] for every budget b
    best_m[b] = 0 for every budget b

    # consider assigning a prefix of length m to individual i
    for each prefix length m from 1 to L:

        # candidate value if m treatment units are assigned to individual i
        for each budget b from m to B:
            cand[b] = dp_prev[b - m] + prefix[i][m]

            # if this improves the best known value at budget b, update it
            if cand[b] > best[b]:
                best[b] = cand[b]
                best_m[b] = m

    # store the chosen prefix length for this individual at each budget
    for each budget b from 0 to B:
        choice[i][b] = best_m[b]

    # move to the next individual
    dp_prev = best

# Step 4: recover allocations by backward pass
# For each budget beta, trace backward through choice to determine
# how many levels were assigned to each individual
for each budget beta from 1 to B:
    remaining_budget = beta

    for individuals i in reverse order:
        m = choice[i][remaining_budget]
        assign levels 1 through m to individual i
        remaining_budget = remaining_budget - m
\end{verbatim}
To recover feasible allocations, the selected prefix length for each individual and budget is stored during the forward pass and used in a backward pass to reconstruct the allocation corresponding to each budget level.

\subsection{Dataset Generation}\label{apd:data}
In the definitions below, $\mu$ is the mean of the per-level benefit draws in Datasets A and B, and $\sigma_s$ is the scale parameter of the log-normal distribution over individual scales in Dataset D. The specific values used in our experiments are set in the data generation scripts available at \url{https://github.com/gilstero/ML-TOC}.

\textbf{Dataset A: Strictly Increasing Marginal Benefits}


\[
z_i \sim \mathcal{N}(\mu \mathbf{1}_L,\ 0.5^2 I_L)
\]

\[
\delta_i^{(1)} = \max\{0,\ z_i^{(1)}\}
\]

\[
\delta_i^{(\ell)} = \max\left\{z_i^{(\ell)},\ \delta_i^{(\ell-1)} + 1\right\}
\text{ for } \ell = 2,\dots,L
\]

\bigskip

\textbf{Dataset B: Random Marginal Benefits}


\[
z_i \sim \mathcal{N}(\mu \mathbf{1}_L,\ 0.5^2 I_L)
\]

\[
\delta_i^{(\ell)} = \max\{0,\ z_i^{(\ell)}\}
\text{ for } \ell = 1,\dots,L
\]

\bigskip

\textbf{Dataset C: Strictly Decreasing Marginal Benefits}

Dataset C is constructed as the mirror image of Dataset A. For each individual $i$, the strictly increasing marginal benefit sequence
$\{\delta_i^{(1)},\dots,\delta_i^{(L)}\}$ generated under Dataset A is reversed to obtain
$\{\delta_i^{(L)},\dots,\delta_i^{(1)}\}$, yielding strictly decreasing marginal benefits across treatment levels.

\textbf{Dataset D: Heterogeneous Strictly Increasing Marginal Benefits}

\[
s_i \sim \text{LogNormal}(0, \sigma_s^2)
\]

\[
z_i^{(\ell)} \sim \mathcal{N}(s_i,\ 0.5^2)
\]

\[
\delta_i^{(1)} = \max\{0, z_i^{(1)}\}
\]

\[
\delta_i^{(\ell)} = \max\left\{ z_i^{(\ell)},\ \delta_i^{(\ell-1)} + 1 \right\} \quad \text{for } \ell = 2,\dots,L
\]

Each individual $i$ is assigned a scale $s_i$, which determines how large their treatment effects are overall. The values $z_i^{(\ell)}$ are noisy draws of the marginal benefit at each level, centered around $s_i$. The first level is truncated to be nonnegative, and each subsequent level is forced to be at least one unit larger than the previous level. This ensures that marginal benefits strictly increase for each individual, while allowing different individuals to have very different magnitudes of treatment effects.

\subsection{Strictly Decreasing Aggregate Plot}\label{apd:SD}

\begin{figure} [H]
    \centering
    \includegraphics[width=1\linewidth]{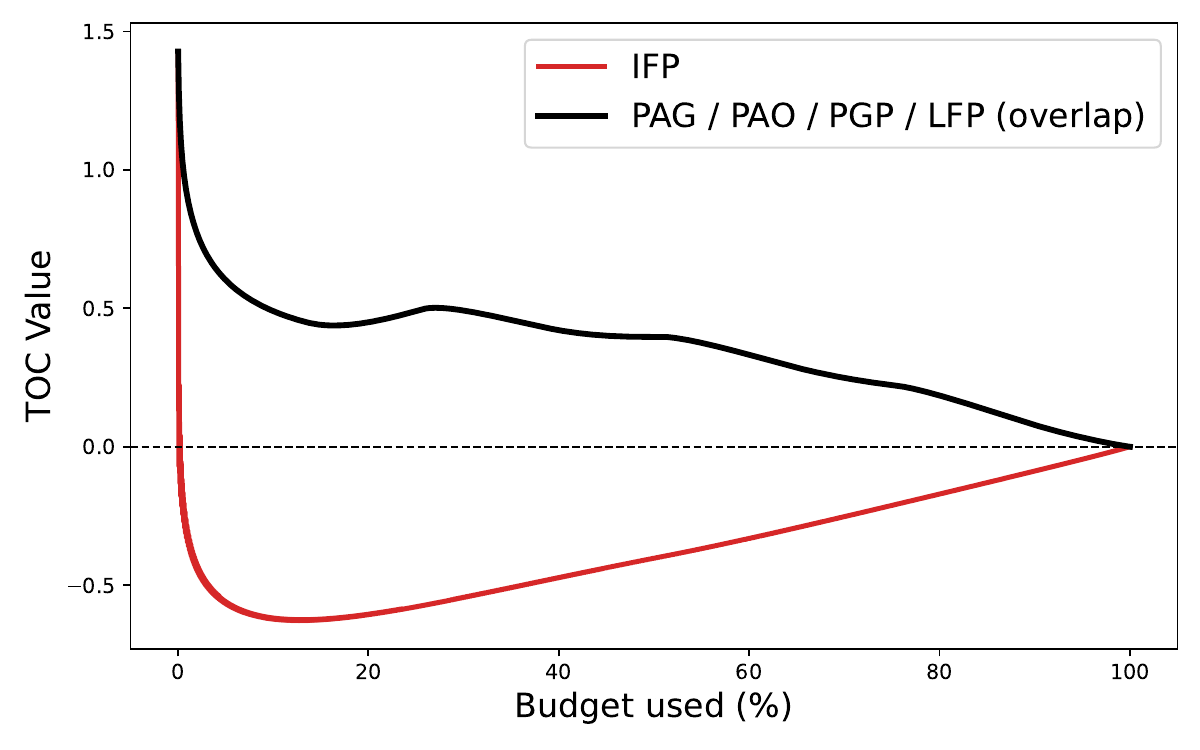}
    \caption{Average TOC curves across budgets for the strictly decreasing marginal benefit dataset. Curves are averaged over 100 datasets and plotted relative to a random baseline. $\pm$ one standard deviation error bars are plotted but too small to be visually distinguishable.}
    \label{fig:7}
\end{figure}

\subsection{Run Time Data}\label{apd:timingchart}

\begin{table}[H]
    \centering
    \caption{Runtime (seconds) of ranking policies as a function of dataset size $n$ on strictly increasing marginal benefit datasets with $L=4$.}
    \label{tab:timing}
    \footnotesize
    \setlength{\tabcolsep}{4pt}
    \begin{tabular}{lccccc}
        \toprule
        $n$ 
        & \textbf{PGP} 
        & \textbf{IFP} 
        & \textbf{LFP} 
        & \textbf{PAG} 
        & \textbf{PAO} \\
        \midrule
       500 & 0.0008 & 0.0009 & 0.0009 & 0.0041 & 0.6721 \\
       1{,}000 & 0.0017 & 0.0017 & 0.0017 & 0.0086 & 2.6607 \\
       2{,}000 & 0.0033 & 0.0033 & 0.0033 & 0.0178 & 10.7443 \\
       4{,}000 & 0.0069 & 0.0069 & 0.0069 & 0.0483 & 43.1270 \\
       8{,}000 & 0.0143 & 0.0135 & 0.0135 & 0.1529 & 176.8498 \\
        \bottomrule
    \end{tabular}
\end{table}

\subsection{Software}\label{apd:software}

To enable the repeatability and efficiency in the experimental phase of this research, we have developed an accompanying software that allows for ML-AUTOC demonstrations available at \url{https://mlautoc.com}. As stated in the docs, the goal of the software was to facilitate the following:
\begin{enumerate}
\setlength{\topsep}{0pt}
\setlength{\itemsep}{0pt}
    \item Choose the number of patients in the dataset.
    \item Select the number of marginal treatment levels each patient has.
    \item Experiment with the cost structure of the experiment.
    \item Configure the underlying structure of marginal treatment effects.
    \item Optionally upload an individual dataset (CSV) and run an experiment.
    \item View resource allocation graphs and ML-AUTOC scores.
    \item Analyze runtime performance (time logged and displayed as a graph).
\end{enumerate}
The platform is implemented using Next.js, and the backend runs locally on the experimenter's computer. Experiments are executed using common Python libraries to ensure full reproducibility. Source code and documentation are available on the project website under \url{https://mlautoc.com/experiments}.

During research we developed several functions that allow us to generate, rank, and visualize controlled treatment allocation scenarios prior to web deployment. These utilities were used to test our theories on flip rankings, precedence enforcement, and cost heterogeneity before web implementation. These functions can be downloaded as files under \url{https://mlautoc.com/downloads}. If the website is inaccessible, all available code can be found on GitHub at \url{https://github.com/gilstero/ML-TOC}.

\end{document}